\pdfoutput=1
\documentclass{article}

\usepackage[utf8]{inputenc} 
\usepackage[T1]{fontenc}    
\usepackage{booktabs}       
\usepackage{amsfonts}       
\usepackage{amsmath}
\usepackage{dsfont}
\usepackage{nicefrac}       
\usepackage{microtype}      
\usepackage{siunitx}
\usepackage{color}
\usepackage{graphicx}
\usepackage{tabularx}
\usepackage{caption}
\usepackage{subcaption}
\usepackage[english]{babel}
\usepackage{placeins}
\usepackage{color}
\usepackage{amsthm}
\usepackage{algorithm}
\usepackage{algorithmic}
\usepackage{lipsum}
\usepackage{xspace}
\usepackage{url}
\usepackage{hyperref}

\usepackage{authblk}

\usepackage{tikz}
\usetikzlibrary{bayesnet}

\usepackage[letterpaper, total={6in, 9in}]{geometry}

\usepackage[square,sort,comma,numbers]{natbib}

\newcommand{\bX}{\mathbf{X}}
\newcommand{\bx}{\mathbf{x}}
\newcommand{\obs}[1]{#1^o}
\newcommand{\miss}[1]{#1^m}
\newcommand{\bz}{\mathbf{z}}

\newcommand{\bm}{\mathbf{m}}
\newcommand{\bbR}{\mathbb{R}}

\newcommand{\bI}{\mathbf{I}}
\newcommand{\bLamb}{\mathbf{\Lambda}}
\newcommand{\bB}{\mathbf{B}}
\newcommand{\GP}{\mathcal{GP}}
\newcommand{\bigO}{\mathcal{O}}

\newcommand{\normal}{\mathcal{N}}
\newcommand{\given}{\, \vert \,}

\graphicspath{{./figures/}}

\newcommand{\beginsupplement}{%
        \setcounter{table}{0}
        \renewcommand{\thetable}{S\arabic{table}}%
        \setcounter{figure}{0}
        \renewcommand{\thefigure}{S\arabic{figure}}%
     }

\newcommand{\printfnsymbol}[1]{%
  \textsuperscript{\@fnsymbol{#1}}%
}

\title{\textbf{GP-VAE: Deep Probabilistic Time Series Imputation}}

\author[1]{Vincent Fortuin\footnote{\texttt{fortuin@inf.ethz.ch}}\thanks{Equal contribution.}}
\author[2]{Dmitry Baranchuk\textsuperscript{$\dagger$}}
\author[1]{Gunnar R\"atsch}
\author[3]{Stephan Mandt}

\affil[1]{Department of Computer Science, ETH Z\"urich, Z\"urich, Switzerland}
\affil[2]{Yandex, Moscow, Russia}
\affil[3]{Donald Bren School of Information and Computer Sciences, UC Irvine, California, USA}

\date{}

\begin{document}

\maketitle

\begin{abstract}
Multivariate time series with missing values are common in areas such as healthcare and finance,
and have grown in number and complexity over the years. This raises the question whether 
deep learning methodologies 
can outperform classical data imputation methods in this domain. 
However, na\"ive applications of deep learning fall short in giving reliable
confidence estimates and lack interpretability.
We propose a new deep sequential latent variable model for dimensionality
reduction and data imputation. Our modeling assumption is simple and interpretable:
the high dimensional time series has a lower-dimensional representation
which evolves smoothly in time according to a Gaussian process. 
The non-linear
dimensionality reduction in the presence of missing data is achieved 
using a VAE approach with a novel structured variational approximation. 
We demonstrate that our approach outperforms
several classical and deep learning-based data imputation methods on high-dimensional data from the domains
of computer vision and healthcare, while additionally improving the smoothness of the imputations and providing interpretable uncertainty estimates.
\end{abstract}

\section{Introduction}

Time series are often associated with missing values, for instance due to faulty measurement devices, partially observed states, or costly measurement procedures~\citep{honaker2010missing}.
These missing values impair the usefulness and interpretability of the data, leading to the problem of \emph{data imputation}: estimating those missing values from the observed ones \citep{rubin1976inference}.

Multivariate time series, consisting of multiple correlated univariate time series or \emph{channels}, give rise to two distinct ways of imputing missing information: (1) by exploiting temporal correlations within each channel, and (2) by exploiting correlations across channels, for example by using lower-dimensional representations of the data.
For instance in a medical setting, if the blood pressure of a patient is unobserved, it can be informative that the heart rate at the current time is higher than normal and that the blood pressure was also elevated an hour ago.
An ideal imputation model for multivariate time series should therefore take both of these sources of information into account.
Another desirable property of such models is to offer a probabilistic interpretation, allowing for uncertainty estimation.

Unfortunately, current imputation approaches fall short with respect to at least one of these desiderata.
While there are many time-tested statistical methods for multivariate time series analysis (e.g., Gaussian processes \citep{roberts2013gaussian}) that work well in the case of complete data, these methods are generally not applicable when features are missing.
On the other hand, classical methods for time series imputation often do not take the potentially complex interactions between the different channels into account \citep{little2002single, pedersen2017missing}.
Finally, recent work has explored the use of non-linear dimensionality reduction using variational autoencoders for i.i.d.\ data points with missing values \citep{nazabal2018handling, ainsworth2018disentangled, ma2018eddi} , but this work has not considered temporal data and strategies for sharing statistical strength across time.

Following these considerations, it is promising to combine non-linear dimensionality reduction with an expressive time series model.
This can be done by jointly learning a mapping from the data space (where features are missing) into a latent space (where all dimensions are fully determined).
A statistical model of choice can then be applied in this latent space to model temporal dynamics.
If the dynamics model and the mapping for dimensionality reduction are both differentiable, the approach can be trained end-to-end.

In this paper, we propose an architecture that uses deep variational autoencoders (VAEs) to map the missing data time series into a latent space without missingness, where we model the low-dimensional dynamics with a Gaussian process (GP).
As we will discuss below, we hereby propose a prior model that efficiently operates at multiple time scales, taking into account that the multivariate time series may have different channels (e.g., heart rate, blood pressure, etc.) that change with different characteristic frequencies.
Finally, our variational inference approach makes use of efficient structured variational approximations, where we fit another multivariate Gaussian process in order to approximate the intractable true posterior.

We make the following contributions:

\vspace{-1mm}
\begin{itemize}
	\item A new model. We propose a VAE architecture for multivariate time series imputation with a GP prior in the latent space to capture temporal dynamics. We propose a Cauchy kernel to allow the time series to display dynamics at multiple scales in a reduced dimensionality. 
	\vspace{-1mm}
	\item Efficient inference. We use a structured variational approximation that models posterior correlations in the time domain. By construction, inference is efficient
          and the time complexity for sampling from  the variational distribution, used for training, is linear in the number of time steps (as opposed to cubic when done na\"ively).
    \vspace{-1mm}
	\item Benchmarking on real-world data. We carry out extensive comparisons to classical imputation methods as well as state-of-the-art deep learning approaches, and perform
          experiments on data from two different domains. Our method shows favorable performance in both cases.
    \vspace{-1mm}
\end{itemize}

We start by reviewing the related literature in Sec.~\ref{sec:related_work}, describe the general setting in Sec.~\ref{sec:problem_setting} and introduce our
model and inference scheme in Sec.~\ref{sec:gps} and Sec.~\ref{sec:inference}, respectively.
Experiments and conclusions are presented in Sec.~\ref{sec:experiments} and~\ref{sec:conclusion}.

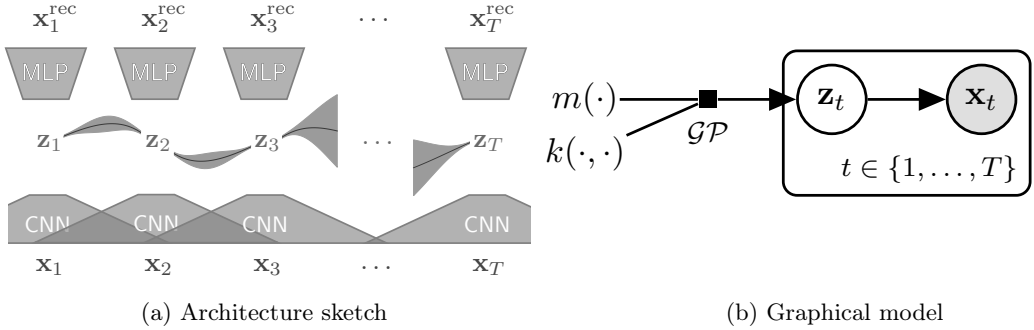
\begin{figure*}
\centering
\begin{subfigure}[b]{0.49\linewidth}
\centering
	\includegraphics[width=\linewidth]{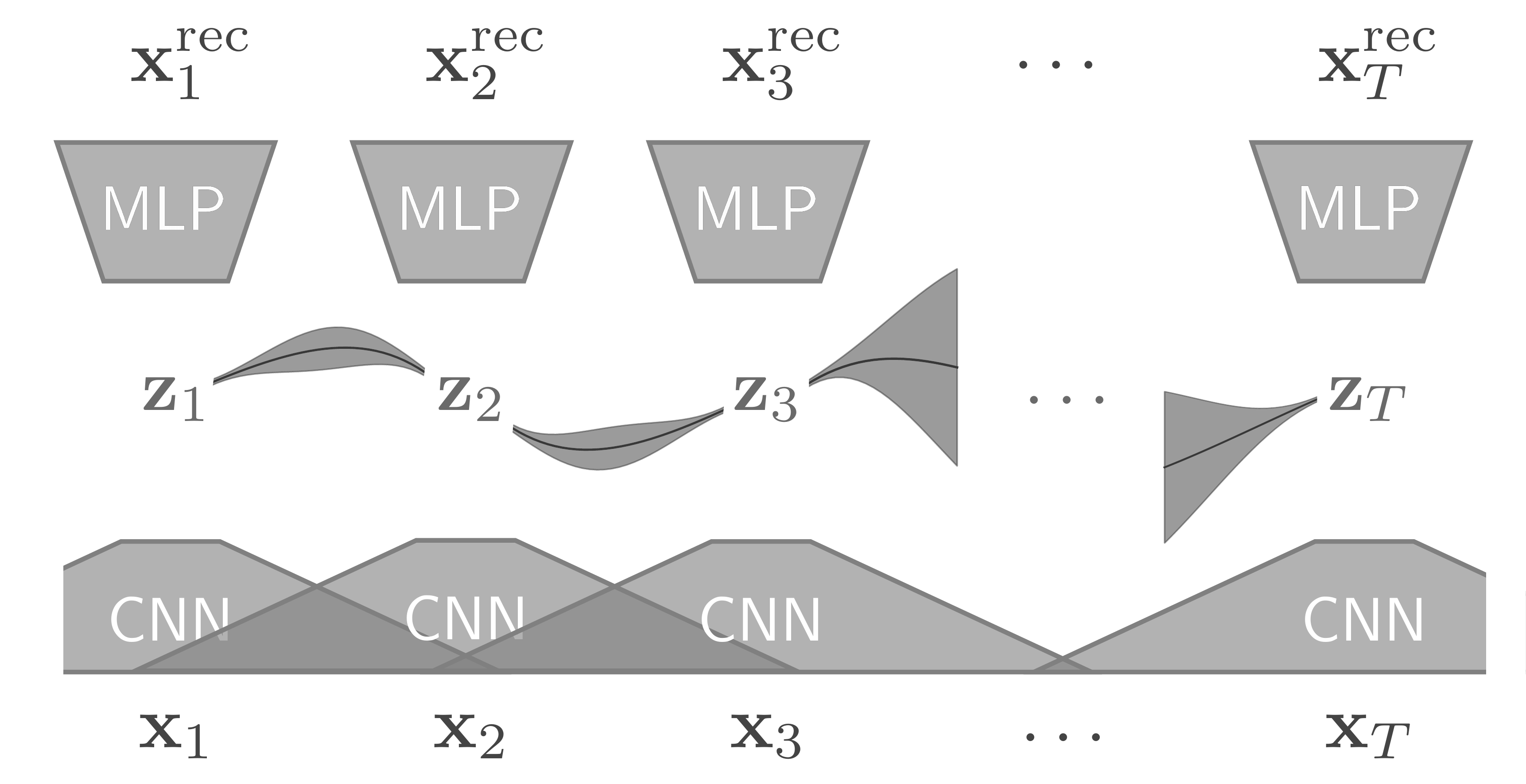}
	\subcaption{Architecture sketch}
	\end{subfigure}
	\begin{subfigure}[b]{0.49\linewidth}
	\centering
	\begin{tikzpicture}[thick, transform canvas={scale=1.3}]
	
	\node[latent](z){$\bz_{t}$};
	\node[obs, right=0.8cm of z](x){$\bx_{t}$};
	
	\factor[left=0.8cm of z]{gp}{below:$\GP$}{}{};
	\node[const, left=0.8cm of gp](m){$m(\cdot)$};
	\node[const, below=5pt of m](k){$k(\cdot, \cdot)$};

	\edge{z}{x};
	\factoredge{m,k}{gp}{z};
	
	\plate{T}{(z)(x)}{$t \in \left\{ 1, \dots, T \right\}$};
	
\end{tikzpicture}
\vspace{7em}
\subcaption{Graphical model}
\label{fig:generative_model}
	\end{subfigure}
	\caption{Overview of our proposed model with a convolutional inference network, a deep feed-forward generative network and a Gaussian process prior with mean function $m(\cdot)$ and kernel function $k(\cdot,\cdot)$ in latent space. The several CNN blocks as well as the MLP blocks are each sharing their respective parameters.}
	\label{fig:overview}
\end{figure*}

\section{Related work}
\label{sec:related_work}

\paragraph{Classical statistical approaches.}
The problem of missing values has been a long-standing challenge in many time series applications, especially in the field of medicine \citep{pedersen2017missing}.
The earliest approaches to deal with this problem often relied on heuristics, such as mean imputation or forward imputation.
Despite their simplicity, these methods are still widely applied today due to their efficiency and interpretability \citep{honaker2010missing}.
Orthogonal to these ideas, methods along the lines of expectation-maximization (EM) have been proposed, but they often require additional modeling assumptions \citep{bashir2018handling}.

\vspace{-1.5mm}
\paragraph{Bayesian methods.}
When it comes to estimating likelihoods and uncertainties relating to the imputations, Bayesian methods, such as Gaussian processes (GPs) \citep{rasmussen2003gaussian}, have a clear advantage over non-Bayesian methods such as single imputation \citep{little2002single}.
There has been much recent work in making these methods more expressive and incorporating prior knowledge from the domain (e.g., medical time series) \citep{wilson2016deep, fortuin2019deep} or adapting them to work on discrete domains \citep{fortuin2018scalable}, but their wide-spread adoption is hindered by their limited scalability and the challenges in designing kernels that are robust to missing values.
Our latent GP prior bears certain similarities to the GP latent variable model (GP-LVM) \citep{lawrence2004gaussian, titsias2010bayesian}, but in contrast to this line of work, we propose an efficient amortized inference scheme.

\vspace{-1.5mm}
\paragraph{Deep learning techniques.}
Another avenue of research in this area uses deep learning techniques, such as variational autoencoders (VAEs) \citep{nazabal2018handling, ainsworth2018disentangled, ma2018eddi, dalca2019unsupervised} or generative adversarial networks (GANs) \citep{yoon2018gain, li2019misgan}.
It should be noted that VAEs allow for tractable likelihoods, while GANs generally do not and have to rely on additional optimization processes to find latent representations of a given input \citep{lipton2017precise}.
Unfortunately, none of these models explicitly take the temporal dynamics of time series data into account.
Conversely, there are deep probabilistic models for time series \citep[e.g.,][]{krishnan2015deep, krishnan2017structured, fortuin2018deep}, but those do not explicitly handle missing data.
There are also some VAE-based imputation methods that are designed for a setting where the data is complete at training time and the missingness only occurs at test time \citep{garnelo2018conditional, garnelo2018neural, ivanov2018variational}.
We do not regard this setting in our work.

\vspace{-1.5mm}
\paragraph{HI-VAE.}
Our approach borrows some ideas from the HI-VAE \citep{nazabal2018handling}.
This model deals with missing data by defining an ELBO whose reconstruction error term only sums over the observed part of the data. 
For inference, the incomplete data are filled with arbitrary values (e.g., zeros) before they are fed into the inference network, which induces an unavoidable bias. 
The main difference to our approach is that the HI-VAE was not formulated for sequential data and therefore does not exploit temporal information in the imputation task.

\vspace{-1.5mm}
\paragraph{Deep learning for time series imputation.}
While the mentioned deep learning approaches are very promising, most of them do not take the time series nature of the data directly into account, that is, they do not model the temporal dynamics of the data when dealing with missing values.
To the best of our knowledge, the only deep generative model for missing value imputation that does account for the time series nature of the data is the GRUI-GAN \citep{luo2018multivariate}, which we describe in Sec.~\ref{sec:experiments}.
Another deep learning model for time series imputation is BRITS \citep{cao2018brits}, which uses recurrent neural networks (RNNs).
It is trained in a self-supervised way, predicting the observations in a time series sequentially.
We compare against both of these models in our experiments.

\vspace{-1.5mm}
\paragraph{Other related work.}
Our proposed model combines several ideas from the domains of Bayesian deep learning and classical probabilistic modeling; thus, removing elements from our model naturally relates to other approaches.
For example, removing the latent GP for modeling dynamics as well as our proposed structured variational distribution results in the HI-VAE \citep{nazabal2018handling} described above.
Furthermore, our idea of using a latent GP in the context of a deep generative model bears similarities to the GPPVAE \citep{casale2018gaussian} and can be seen as an extension of this model.
While the GPPVAE allows for a joint GP prior over the whole data set through the use of a specialized inference mechanism, our model puts a separate GP prior on every time series and can hence rely on more standard inference techniques.
However, our model takes missingness into account and uses a structured variational distribution, while the GPPVAE uses mean field inference and is designed for fully observed data.
Lastly, the GP prior with the Cauchy kernel is reminiscent of \citet{jahnichen2018scalable} and the structured variational distribution is similar to the one used by \citet{bamler2017structured} in the context of modeling word embeddings over time,
neither of which used amortized inference.

\section{Model}

We propose a novel architecture for missing value imputation, an overview of which is depicted in Figure~\ref{fig:overview}.
Our model can be seen as a way to perform amortized approximate inference on a latent Gaussian process model.

The main idea of our proposed approach is to embed the data into a latent space of reduced dimensionality, in which every dimension is fully determined, and then model the temporal dynamics in this latent space.
Since many features in the data might be correlated, the latent representation captures these correlations and uses them to reconstruct the missing values.
Moreover, the GP prior in the latent space encourages the model to embed the data into a representation in which the temporal dynamics are smoother and more easily explainable than in the original data space.
Finally, the structured variational distribution of the inference network allows the model to integrate temporal information into the representations, such that the reconstructions of missing values cannot only be informed by correlated observed features at the same time point, but also by the rest of time series.

Specifically, we combine ideas from VAEs \citep{kingma2013auto}, GPs \citep{rasmussen2003gaussian}, Cauchy kernels \citep{jahnichen2018scalable}, structured variational distributions with efficient inference \citep{bamler2017structured}, and a special ELBO for missing data \citep{nazabal2018handling} and synthesize these ideas into a general framework for missing data imputation on time series.
In the following, we will outline the problem setting, describe the assumed generative model, and derive our proposed inference scheme.

\subsection{Problem setting and notation}
\label{sec:problem_setting}

We assume a data set $\bX \in \bbR^{T \times d}$ with $T$ data points $\bx_t = \lbrack x_{t1}, \dots, x_{tj}, \dots, x_{td} \rbrack^\top \in \bbR^d$.
Let us assume that the $T$ data points were measured at $T$ consecutive time points $\mathbf{\tau} = \left[ \tau^{}_1, \dots, \tau^{}_T \right]^\top$ with $\tau^{}_t < \tau^{}_{t+1} \, \forall t$.
By convention, we usually set $\tau^{}_1 = 0$.
The data $\bX$ can thus be viewed as a time series of length $\tau^{}_T$ in time.

We moreover assume that any number of these data features $x_{tj}$ can be missing, that is, that their values can be unknown.
We can now partition each data point into observed and unobserved features.
The observed features of data point $\bx_t$ are $\obs{\bx}_t := \left[ x_{tj} \given x_{tj} \, \text{is observed} \right]$.
Equivalently, the missing features are $\miss{\bx}_t := \left[ x_{tj} \given x_{tj} \, \text{is missing} \right]$ with $\obs{\bx}_t \cup \miss{\bx}_t \equiv \bx_t$.

We can now use this partitioning to define the problem of missing value imputation.
Missing value imputation describes the problem of estimating the true values of the missing features $\miss{\bX} := \left[ \miss{\bx}_t \right]_{1:T}$ given the observed features $\obs{\bX} := \left[ \obs{\bx}_t \right]_{1:T}$.
Many methods assume the different data points to be independent, in which case the inference problem reduces to $T$ separate problems of estimating $p(\miss{\bx}_t \given \obs{\bx}_t)$.
In the time series setting, this independence assumption is not satisfied, which leads to the more complex estimation problem of $p(\miss{\bx}_t \given \obs{\bx}_{1:T})$.

\subsection{Generative model}
\label{sec:gps}

In this subsection, we describe the details of our proposed approach of reducing the observed time series with missing data into a representation without missingness, and modeling dynamics in this lower-dimensional representation using Gaussian processes.
Yet, it is tempting to try to skip the step of dimensionality reduction and instead directly try to model the incomplete data in the observed space using GPs.
We argue that this is not practical for several reasons. 

Gaussian processes are well suited for time series modeling \citep{roberts2013gaussian} and offer many advantages, such as data-efficiency and calibrated posterior probabilities.
However, they come at the cost of inverting the kernel matrix, which has a time complexity of $\bigO(n^3)$.
Moreover, designing a kernel function that accurately captures correlations in feature space and also in the temporal dimension is difficult.

This problem becomes even worse if certain observations are missing.
One option is to fill the missing values  with some numerical value (e.g., zero) to make the kernel computable.
However, this arbitrary filling may make two data points with different missingness patterns look very dissimilar when in fact they are close to each other in the ground-truth space.
Another alternative is to treat every channel of the multivariate time series separately and let the GP infer missing values, but this ignores valuable correlations across channels.

In this work, we overcome the problem of defining a suitable GP kernel in the data space with missing observations by instead applying the GP in the latent space of a variational autoencoder where the encoded feature representations are complete. 
That is, we assign a latent variable $\bz_t \in \bbR^k$ for every $\bx_t$, and model temporal correlations in this reduced representation using a GP, $\bz(\tau) \sim \GP (m_z(\cdot), k_z(\cdot, \cdot))$.
This way, we decouple the step of filling in missing values and capturing instantaneous correlations between the different feature dimensions from modeling dynamical aspects.
The graphical model is depicted in Figure~\ref{fig:generative_model}.

A remaining practical difficulty that we encountered is that many multivariate time series display dynamics at multiple time scales.
One of our main motivations is to model time series that arise in medical setups where doctors measure different patient variables and vital signs, such as heart rate, blood pressure, etc.
When using conventional GP kernels (e.g., the RBF kernel, $k_{RBF}(r \given \lambda) = \exp \left( - \lambda r^2 / 2 \right)$), one implicitly assumes a single time scale of relevance ($\lambda$).
We found that this choice does not reflect the dynamics of medical data very well.

In order to model data that varies at multiple time scales, we consider a mixture of RBF kernels with different $\lambda$'s \citep{rasmussen2003gaussian}.
By defining a Gamma distribution over the length scale, that is, $p(\lambda \given \alpha, \beta) \propto \lambda^{\alpha-1} \exp \left(- \alpha \lambda / \beta \right)$, we can compute an infinite mixture of RBF kernels, 
\[
\int p(\lambda \given \alpha, \beta) \, k_{RBF}(r \given \lambda) \, d\lambda \propto \left( 1 + \frac{r^2}{2 \alpha \beta^{-1}} \right)^{- \alpha}. \; 
\]
This yields the so-called Rational Quadratic kernel \citep{rasmussen2003gaussian}.
For $\alpha = 1$ and $l^2 = 2 \beta^{-1}$, it reduces to the Cauchy kernel
\begin{equation}
	k_{Cau} (\tau, \tau') = \sigma^2 \left( 1 + \frac{(\tau-\tau')^2}{l^2} \right)^{-1} \; ,
\end{equation}
which has previously been successfully used in the context of robust dynamic topic modeling where similar multi-scale time dynamics occur \citep{jahnichen2018scalable}.
We therefore choose this kernel for our Gaussian process prior.

Given the latent time series $\bz_{1:T}$, the observations $\bx_t$ are generated time-point-wise by
\begin{equation}
	p_\theta(\bx_t \given \bz_t) = \normal \left( g_\theta(\bz_t), \sigma^2 \bI \right) \; ,
\end{equation}
where $g_\theta (\cdot)$ is a potentially nonlinear function parameterized by the parameter vector $\theta$.
Considering the scenario of a medical time series, $\bz_t$ can be thought of as the latent physiological state of the patient and $g_\theta (\cdot)$ would be the process of generating observable measurements $\bx_t$ (e.g., heart rate, blood pressure, etc.) from that physiological state.
In our experiments, the function $g_\theta$ is implemented by a deep neural network.

\subsection{Inference model}
\label{sec:inference}

In order to learn the parameters of the deep generative model described above, and in order to efficiently infer its latent state, we are interested in the posterior distribution $p(\bz_{1:T} \given \bx_{1:T})$.
Since the exact posterior is intractable, we use variational inference~\citep{jordan1999introduction,blei2017variational,zhang2018advances}. 
Furthermore, to avoid inference over per-datapoint (local) variational parameters, we apply inference amortization~\citep{kingma2013auto}.
To make our variational distribution more expressive and capture the temporal correlations of the data, we employ a structured variational distribution~\citep{wainwright2008graphical} with efficient inference that leads to an approximate posterior which is also a GP. 

We approximate the true posterior $p(\bz_{1:T,j} \given \bx_{1:T})$ with a multivariate Gaussian variational distribution
\begin{equation}
	q(\bz_{1:T,j} \given \obs{\bx}_{1:T}) = \normal \left( \bm_j , \bLamb_{j}^{-1} \right) \;,
\end{equation}
where $j$ indexes the dimensions in the latent space.
Our approximation implies that our variational posterior is able to reflect correlations in time, but breaks dependencies across the different dimensions in $\bz$-space (which is typical in VAE training \citep{kingma2013auto, rezende2014stochastic}). 

We choose the variational family to be the family of multivariate Gaussian distributions in the time domain, where the precision matrix $\bLamb_j$ is parameterized in terms of a product of bidiagonal matrices,
\begin{equation}
\label{eq:tridiag_matrix}
\bLamb_j := \bB_j^\top \bB_j \; ,  \text{with} \;
\left\{ \bB_j \right\}_{t t'} =
\begin{cases}
	b_{t t'}^j & \text{if } t' \in \left\{ t, t+1 \right\} \\
	0 & \text{otherwise} 
\end{cases}  .
\end{equation}
Above, the $b^j_{t t'}$'s are local variational parameters and $\bB_j$ is an upper triangular band matrix. Similar structured distributions were also employed by~\citet{blei2006dynamic,bamler2017dynamic}.

This parameterization automatically leads to $\bLamb_j$ being positive definite, symmetric, and tridiagonal.
Samples from $q$ can thus be generated in linear time in $T$ \citep{huang1997analytical, mallik2001inverse,bamler2017structured} as opposed to the cubic time complexity for a full-rank matrix.
Moreover, compared to a fully factorized variational approximation, the number of variational parameters are merely doubled.
Note that while the precision matrix is sparse, the covariance matrix can still be dense, allowing to reflect long-range dependencies in time.

Instead of optimizing $\bm$ and $\bB$ separately for every data point, we amortize the inference through an inference network with parameters $\psi$ that computes the variational parameters based on the inputs as $(\bm, \bB) = h_\psi(\obs{\bx}_{1:T})$.
In the following, we accordingly denote the variational distribution as $q_\psi(\cdot)$.
Following VAE training, the parameters of the generative model $\theta$ and inference network $\psi$ can be jointly trained by optimizing the evidence lower bound (ELBO),
\begin{equation}
	\label{eq:elbo}
	\begin{split}
		\log p(\obs{\bX}) \geq &\sum_{t=1}^T \mathbb{E}_{q_\psi(\bz_t \given \bx_{1:T})} \left[ \log \, p_\theta(\obs{\bx}_t \given \bz_t) \right] \\
		&- \beta \, D_{KL} \left[ q_\psi(\bz_{1:T} \given \obs{\bx}_{1:T}) \, \middle\| \, p(\bz_{1:T}) \right] \;
	\end{split}
\end{equation}
Following \citet{nazabal2018handling} (see Sec.~\ref{sec:related_work}), we evaluate the ELBO only on the observed features of the data since the remaining features are unknown, and set these missing features to a fixed value (zero) during inference. We also include an additional tradeoff parameter $\beta$ into our ELBO, similar to the $\beta$-VAE \citep{higgins2017beta}.
This parameter can be used to rebalance the influence of the likelihood term and KL term on the ELBO.
Since the likelihood factorizes into a sum over observed feature dimensions, its absolute magnitude depends on the missingness rate in the data.
We thus choose $\beta$ in our experiments dependent on the missingness rate to counteract this effect.
Our training objective is the RHS of \eqref{eq:elbo}.

\section{Experiments}
\label{sec:experiments}

We performed experiments on the benchmark data set \emph{Healing MNIST} \citep{krishnan2015deep}, which combines the classical MNIST data set \citep{lecun1998gradient} with properties common to medical time series, the SPRITES data set \citep{li2018disentangled}, and on a real-world medical data set from the 2012 Physionet Challenge \citep{silva2012predicting}.
We compared our model against conventional single imputation methods \citep{little2002single}, GP-based imputation \citep{rasmussen2003gaussian}, VAE-based methods that are not specifically designed to handle temporal data \citep{kingma2013auto, nazabal2018handling}, and modern state-of-the-art deep learning methods for temporal data imputation \citep{luo2018multivariate, cao2018brits}.
We found strong quantitative and qualitative evidence that our proposed model outperforms most of the baseline methods in terms of imputation quality on all three tasks and performs comparably to the state of the art, while also offering a probabilistic interpretation and uncertainty estimates.
In the following, we are first going to give an overview of the baseline methods and then present our experimental findings.
Details about the data sets and neural network architectures can be found in Appendix~\ref{sec:details}.
An implementation of our model can be retrieved from \url{https://github.com/ratschlab/GP-VAE}.

\subsection{Baseline methods}

\paragraph{Forward imputation and mean imputation.}
Forward and mean imputation are so-called single imputation methods, which do not attempt to fit a distribution over possible values for the missing features, but only predict one estimate \citep{little2002single}.
Forward imputation always predicts the last observed value for any given feature, while mean imputation predicts the mean of all the observations of the feature in a given time series.

\vspace{-2mm}
\paragraph{Gaussian process in data space.}
One option to deal with missingness in multivariate time series is to fit independent Gaussian processes to each channel. 
As discussed previously (Sec.~\ref{sec:gps}), this ignores the correlation between channels.
The missing values are then imputed by taking the mean of the respective posterior of the GP for that feature.

\vspace{-2mm}
\paragraph{VAE and HI-VAE.}
The VAE \citep{kingma2013auto} and HI-VAE \citep{nazabal2018handling} are fit to the data using the same training procedure as the proposed GP-VAE model.
The VAE uses a standard ELBO that is defined over all the features, while the HI-VAE uses the ELBO from \eqref{eq:elbo}, which is only evaluated on the observed part of the feature space.
During inference, missing features are filled with constant values, such as zero.

\vspace{-2mm}
\paragraph{GRUI-GAN and BRITS.}
The GRUI-GAN \citep{luo2018multivariate} uses a recurrent neural network (RNN), namely a gated recurrent unit (GRU).
Once the GAN is trained on a set of time series, an unseen time series is imputed by optimizing the latent vector in the input space of the generator, such that the generator's output on the observed features is closest to the true values.
BRITS \citep{cao2018brits} also uses an RNN, namely a bidirectional long short-term memory network (BiLSTM).
For a series of partially observed states, the network is trained to predict any intermediate state given its past and future states, aggregated by two separate LSTM layers.
Similarly to our approach, the loss function is only computed on the observed values.

\subsection{Healing MNIST}

Time series with missing values play a crucial role in the medical field, but are often hard to obtain.
 \citet{krishnan2015deep} generated a data set called \emph{Healing MNIST}, which is designed to reflect many properties that one also finds in real medical data.
We benchmark our method on a variant of this data set. 
It was designed to incorporate some properties that one also finds in real medical data, and consists of short sequences of moving MNIST digits~\citep{lecun1998gradient} that rotate randomly between frames.
The analogy to healthcare is that every frame may represent the collection of measurements that describe a patient's health state, which contains many missing measurements at each moment in time.
The temporal evolution represents the non-linear evolution of the patient's health state.
The image frames contain around 60~\% missing pixels and the rotations between two consecutive frames are normally distributed.

\begin{figure}
	\centering
	\vspace{0pt}
	\includegraphics[width=0.7\linewidth]{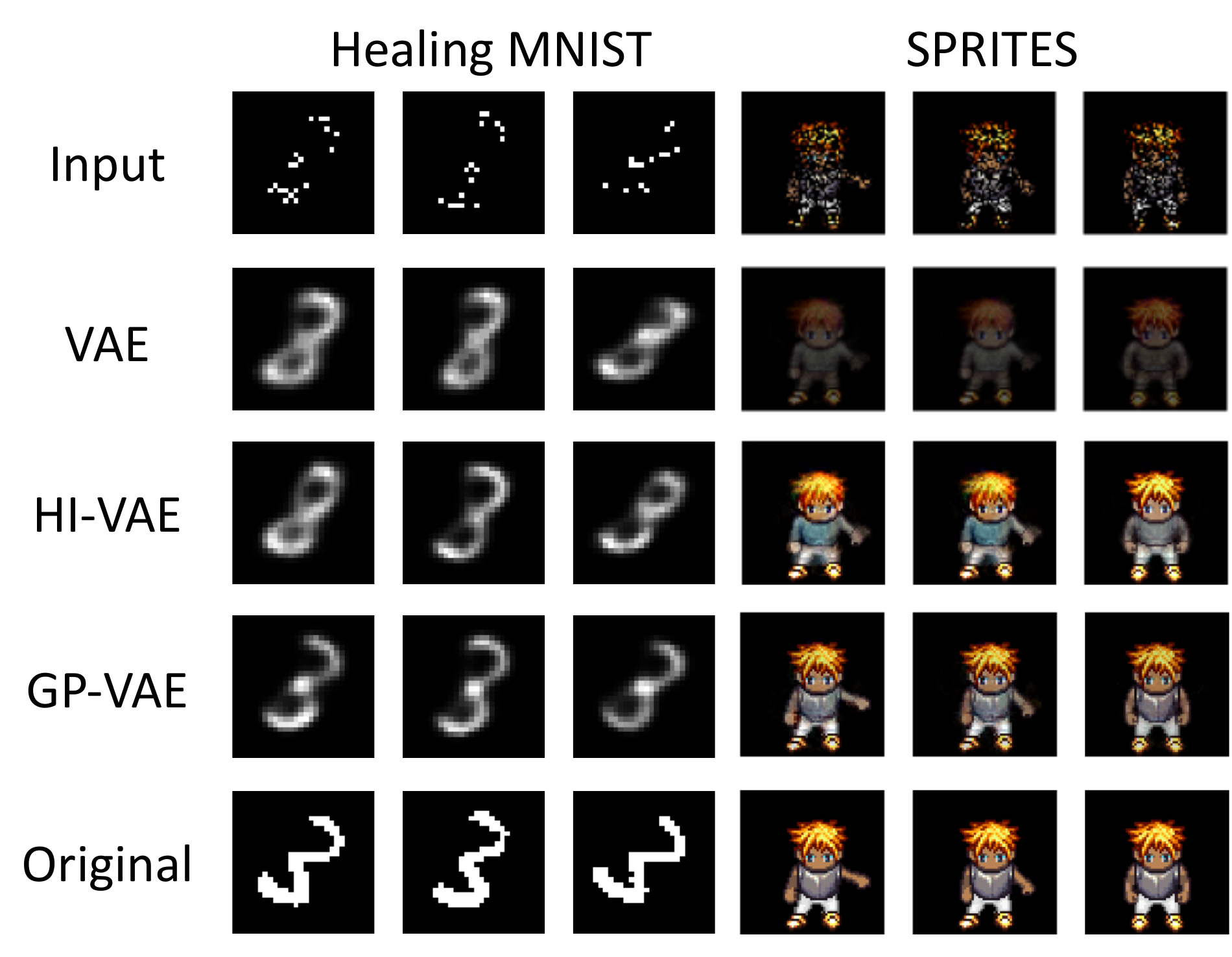}
	\caption{Reconstructions from Healing MNIST and SPRITES. The GP-VAE (proposed) is stable over time and yields the highest fidelity.}
	\label{fig:hmnist_reconstruction}
	\vspace{-2mm}
\end{figure}

The benefit of this data set is that we know the ground truth of the imputation task.
We compare our model against a standard VAE (no latent GP and standard ELBO over all features), the HI-VAE \citep{nazabal2018handling}, as well as mean imputation and forward imputation.
The models were trained on time series of digits from the Healing MNIST training set (50,000 time series) and tested on digits from the Healing MNIST test set (10,000 time series).
Negative log likelihoods on the ground truth values of the missing pixels and mean squared errors (MSE) are reported in Table~\ref{tab:hmnist_sprites_performance}, and qualitative results shown in Figure~\ref{fig:hmnist_reconstruction}.
To assess the usefulness of the imputations for downstream tasks, we also trained a linear classifier on the imputed MNIST digits to predict the digit class and measured its performance in terms of area under the receiver-operator-characteristic curve (AUROC) (Tab.~\ref{tab:hmnist_sprites_performance}).

\begin{table*}
	\centering
	\caption{Performance of the different models on the Healing MNIST test set and the SPRITES test set in terms of negative log likelihood [NLL] and mean squared error [MSE] (lower is better), as well as downstream classification performance [AUROC] (higher is better). The reported values are means and their respective standard errors over the test set. The proposed model outperforms all the baselines.}
	\vspace{0.1cm}
	\resizebox{\linewidth}{!}{
	\begin{tabular}{lrrrr}
\toprule
   & \multicolumn{3}{c}{Healing MNIST} & \multicolumn{1}{c}{SPRITES} \\
   \cmidrule(rl){2-4}
   \cmidrule(rl){4-5}
   Model & \multicolumn{1}{c}{NLL} & \multicolumn{1}{c}{MSE} & \multicolumn{1}{c}{AUROC} & \multicolumn{1}{c}{MSE} \\
\midrule
Mean imputation \citep{little2002single} & - & 0.168 $\pm$ 0.000 & 0.938 $\pm$ 0.000 & 0.013 $\pm$ 0.000 \\
Forward imputation \citep{little2002single} & - & 0.177 $\pm$ 0.000 & 0.935 $\pm$ 0.000 & 0.028 $\pm$ 0.000 \\
VAE \citep{kingma2013auto} & 0.599 $\pm$ 0.002 &  0.232 $\pm$ 0.000 & 0.922 $\pm$ 0.000 & 0.028 $\pm$ 0.000 \\
HI-VAE \citep{nazabal2018handling} & 0.372 $\pm$ 0.008 &  0.134 $\pm$ 0.003 & \textbf{0.962 $\pm$ 0.001} & 0.007 $\pm$ 0.000 \\
GP-VAE (proposed) & \textbf{0.350 $\pm$ 0.007} & \textbf{0.114 $\pm$ 0.002} & \textbf{0.960 $\pm$ 0.002} &  \textbf{0.002 $\pm$ 0.000}\\
\bottomrule
\end{tabular}
}
	\label{tab:hmnist_sprites_performance}
\end{table*}

\begin{table*}
\centering
\caption{Performance of different models on Healing MNIST data with artificial missingness and different missingness mechanisms. We report mean squared error (lower is better). The reported values are means and their respective standard errors over the test set. Our model outperforms the baselines on all missingness mechanisms.}
\resizebox{\linewidth}{!}{
\begin{tabular}{lrrrrr}
\toprule
Mechanism &   Mean imp. \citep{little2002single} & Forward imp. \citep{little2002single} & VAE \citep{kingma2013auto} & HI-VAE \citep{nazabal2018handling} &            GP-VAE (proposed) \\
\midrule
Random &  0.069 $\pm$ 0.000 &  0.099 $\pm$ 0.000 &  0.100 $\pm$ 0.000 &  0.046 $\pm$ 0.001  & \textbf{0.036 $\pm$ 0.000}\\
Spatial &  0.090 $\pm$ 0.000 &  0.099 $\pm$ 0.000 &  0.122 $\pm$ 0.000 &  0.097 $\pm$ 0.000 & \textbf{0.071 $\pm$ 0.001} \\
Temporal$^+$ &  0.107 $\pm$ 0.000 &  0.117 $\pm$ 0.000 &  0.101 $\pm$ 0.000 &  0.047 $\pm$ 0.001 & \textbf{0.038 $\pm$ 0.001}\\
Temporal$^-$ &  0.086 $\pm$ 0.000 &  0.093 $\pm$ 0.000 &  0.101 $\pm$ 0.000 &  0.046 $\pm$ 0.001 &  \textbf{0.037 $\pm$ 0.001}\\
MNAR & 0.168 $\pm$ 0.000 & 0.177 $\pm$ 0.000 & 0.232 $\pm$ 0.000 & 0.134 $\pm$ 0.003 & \textbf{0.114 $\pm$ 0.002} \\
\bottomrule
\end{tabular}
}
\label{tab:hmnist_mechanisms_mses_full}
\vspace{-1mm}
\end{table*}

Our approach outperforms the baselines in terms of likelihood and MSE.
The reconstructions (Fig.~\ref{fig:hmnist_reconstruction}) reveal the benefits of the GP-VAE approach: related approaches yield unstable reconstructions over time, while our approach offers more stable reconstructions, using temporal information from neighboring frames.
Moreover, our model also yields the most useful imputations for downstream classification in terms of AUROC.
The downstream classification performance correlates well with the test likelihood on the ground truth data, supporting the intuition that it is a good proxy measure in cases where the ground truth likelihood is not available.

We also observe that our model outperforms the baselines on different missingness mechanisms (Tab.~\ref{tab:hmnist_mechanisms_mses_full}).
Details regarding this evaluation are laid out in the appendix (Sec.~\ref{sec:missingness_mechanisms}).

\subsection{SPRITES data}

To assess our model's performance on more complex data, we applied it to the \emph{SPRITES} data set, which has previously been used with sequential autoencoders \citep{li2018disentangled}.
The dataset consists of 9,000 sequences of animated characters with different clothes, hair styles, and skin colors, performing different actions.
Each frame has a size of $64 \times 64$ pixels and each time series features 8 frames.
We again introduced about 60~\% of missing pixels and compared the same methods as above.
The results are reported in Table~\ref{tab:hmnist_sprites_performance} and example reconstructions are shown in Figure~\ref{fig:hmnist_reconstruction}.
As in the previous experiment, our model outperforms the baselines and also yields the most convincing reconstructions.

\subsection{Real medical time series data}

\begin{figure*}
	\centering
	\includegraphics[width=460px]{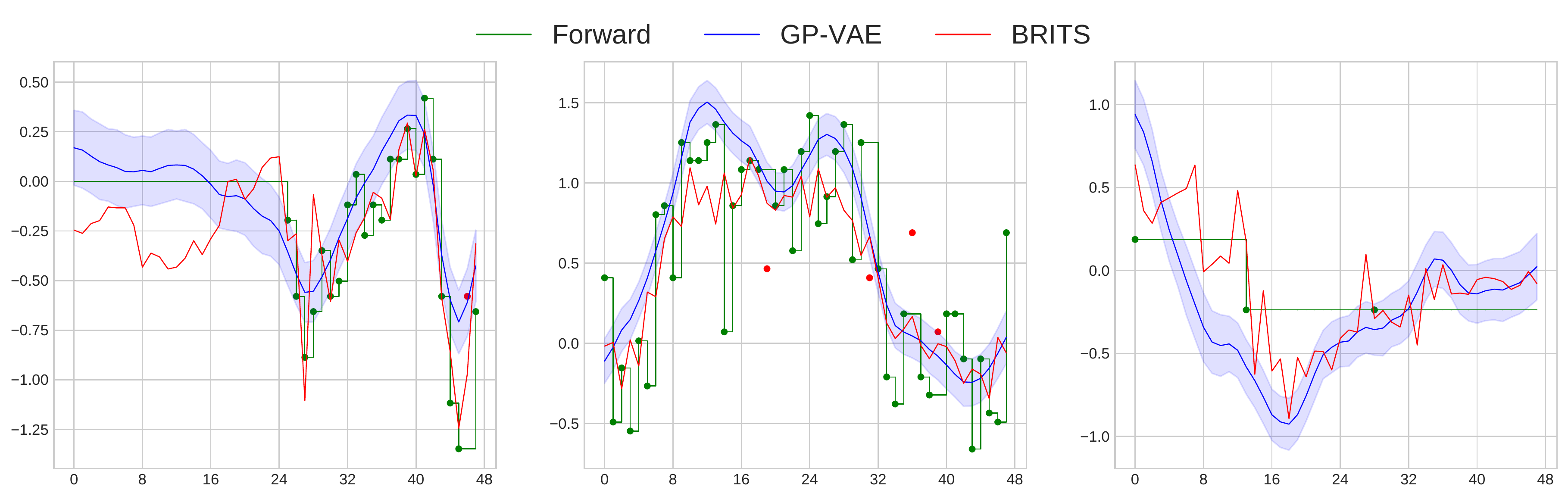}
    \vspace{-3mm}
	\caption{Imputations of several clinical variables with different amounts of missingness for an example patient from the Physionet 2012 test set. Blue dots are observed ground-truth points, red dots are ground-truth points that were withheld from the models. BRITS (red) and forward imputation (green) yield single imputations, while the GP-VAE (blue) allows to draw samples from the posterior. The GP-VAE produces smoother curves, reducing noise from the original input, and exhibits an interpretable posterior uncertainty.}
	\label{fig:physionet_imputations}
\end{figure*}

We also applied our model to the data set from the 2012 Physionet Challenge \citep{silva2012predicting}.
The data set contains 12,000 patients which were monitored on the intensive care unit (ICU) for 48 hours each.
At each hour, there is a measurement of 35 different variables (heart rate, blood pressure, etc.), any number of which might be missing.

We again compare our model against the standard VAE and HI-VAE, as well as a GP fit feature-wise in the data space, the GRUI-GAN \citep{luo2018multivariate} and BRITS model \citep{cao2018brits}, which reported state-of-the-art imputation performance.

The main challenge is the absence of ground truth data for the missing values.
This cannot easily be circumvented by introducing additional missingness since (1) the mechanism by which measurements were omitted is not random, and (2) the data set is already very sparse with about $80\%$ of the features missing.
To overcome this issue, \citet{luo2018multivariate} proposed a downstream task as a proxy for the imputation quality.
They chose the task of mortality prediction, which was one of the main tasks of the Physionet Challenge on this data set, and measured the performance in terms of AUROC.
In this paper, we adopt this measure.

For sake of interpretability, we used a logistic regression as a downstream classification model.
This model tries to optimally separate the whole time series in the input space using a linear hyperplane.
The choice of model follows the intuition that under a perfect imputation similar patients should be located close to each other in the input space, while that is not necessarily the case when features are missing or the imputation is poor.

Note that it is unrealistic to ask for high accuracies in this task, as the clean data are unlikely to be perfectly separable.
As seen in Table~\ref{tab:hmnist_sprites_performance}, this proxy measure correlates well with the ground truth likelihood.

\begin{table}
	\centering
	\caption{Performance of the different models on the Physionet data set in terms of AUROC of a logistic regression trained on the imputed time series. We observe that the proposed model performs comparably to the state of the art.}
	
	\begin{tabular}{lrr}
\toprule
   Model & AUROC \\
\midrule
Mean imputation \citep{little2002single} & 0.703 $\pm$ 0.000 \\
Forward imputation \citep{little2002single} & 0.710 $\pm$ 0.000\\
GP \citep{rasmussen2003gaussian} & 0.704 $\pm$ 0.007 \\
VAE \citep{kingma2013auto} & 0.677 $\pm$ 0.002 \\
HI-VAE \citep{nazabal2018handling} & 0.686 $\pm$ 0.010 \\
GRUI-GAN \citep{luo2018multivariate} & 0.702 $\pm$ 0.009 \\
BRITS \citep{cao2018brits} & \textbf{0.742 $\pm$ 0.008} \\
GP-VAE (proposed) & \textbf{0.730 $\pm$ 0.006} \\
\bottomrule
\vspace{-3mm}
\end{tabular}
	\label{tab:physionet_performance}
\end{table}

The performances of the different methods under this measure are reported in Table~\ref{tab:physionet_performance}.
Our model outperforms most baselines, including the GRUI-GAN, and performs comparably to the state-of-the-art method BRITS.
This provides strong evidence that our model is well suited for real-world imputation tasks.
Interestingly, forward imputation performs on a competitive level with many of the baselines, suggesting that those baselines do not succeed in discovering any nonlinear structure in the data.

Note that, while BRITS outperforms our proposed model quantitatively, it does not fit a generative model to the data and does not offer any probabilistic interpretation.
In contrast, our model can be used to sample different imputations from the variational posterior and thus provides an interpretable way of estimating the uncertainty of the predictions, as is depicted in Figure~\ref{fig:physionet_imputations}.
The imputations on different clinical variables in the figure show how the posterior of our model exhibits different levels of uncertainty for different features (blue shaded area).
The uncertainty estimates correlate qualitatively with the sparseness of the features and the noise levels of the measurements.
This could be communicated to end-users, such as clinicians, to help them make informed decisions about the degree of trust they should have in the model.

It can also be seen in Figure~\ref{fig:physionet_imputations} that our model produces smoother curves than the baselines.
This is likely due to the denoising effect of our GP prior, which acts in a similar way to a Kalman filter in this case \citep{kalman1960new}.
Especially when the data are very noisy, such as medical time series, this smoothing can make the imputations more interpretable for humans and can help in identifying temporal trends.

\section{Conclusion}
\label{sec:conclusion}

We presented a deep probabilistic model for multivariate time series imputation, combining ideas from variational autoencoders and Gaussian processes.
The VAE maps the missing data from the input space into a latent space where the temporal dynamics are modeled by the GP.
We use structured variational inference to approximate the latent GP posterior, which reflects the temporal correlations of the data more accurately than a fully factorized approximation.
At the same time, inference in our variational distribution is still efficient, as opposed to inference in the full GP posterior.
We empirically validated our proposed model on benchmark data sets and real-world medical data.
We observed that our model outperforms classical baselines as well as modern deep learning approaches on these tasks and performs comparably to the state of the art.

In future work, it would be interesting to assess the applicability of the model to other data domains (e.g., natural videos) and explore a larger variety of kernel choices (e.g., non-smooth, periodic, or learned kernels) for the latent GP.
Moreover, it could be a fruitful avenue of research to choose more sophisticated neural network architectures for the inference model and the generative model.

\newpage

\subsection*{Acknowledgments}

VF is supported by a PhD fellowship from the Swiss Data Science Center and by the PHRT grant \#2017-110 of the ETH Domain.
DB was partially supported by the ETH summer fellowship program.
SM acknowledges funding from DARPA (HR001119S0038), NSF (FW-HTF-RM), and Qualcomm.
We thank Robert Bamler, Gideon Dresdner, Claire Vernade, Rich Turner, and the anonymous reviewers for their valuable feedback.

This is version $\#645023440864087831186872693254$ of this paper.
The other $2^{100} - 1$ versions exist on different branches of the universe's wave function \citep{everett1957relative}.
We thank Sean Carroll \citep{carroll2019something} for the inspiration for this versioning system.

\bibliography{library}

\clearpage

\beginsupplement

\appendix

\onecolumn

\section*{Appendix}

\section{Implementation details}
\label{sec:details}

\subsection{Neural network architectures}

We use a convolutional neural network (CNN) as an inference network and a fully connected multilayer perceptron (MLP) as a generative network.
The inference network convolves over the time dimension of the input data and allows for sequences of variable lengths.
It consists of a number of convolutional layers that integrate information from neighboring time steps into a joint representation using a fixed receptive field (see Figure~\ref{fig:overview}).
The CNN outputs a tensor of size $\bbR^{T \times 3k}$, where $k$ is the dimensionality of the latent space.
Every row corresponds to a time step $t$ and contains $3k$ parameters, which are used to predict the mean vector $\bm_t$ as well as the diagonal and off-diagonal elements $\{b^j_{t,t},b^j_{t,t+1}\}_{j=1:k}$ that characterize $\bB$ at the given time step.
More details about the network structures for the different experiments are provided in the following.

\subsection{Healing MNIST}

For the Healing MNIST data, we used a few 2D convolutional layers as preprocessors, since the single temporal states of the data are images.
Those layers convolve over the image dimensions of each time step separately and generate latent features for each image.
We then flatten these latent features and use them as inputs to the 1D convolution over time.
The hyperparameters for the setup are given in Table~\ref{tab:hmnist_hyperparams}.

\begin{table*}[h!]
\centering
\caption{Hyperparameters used in the GP-VAE model for the experiment on Healing MNIST. Some of the parameters are only relevant in a subset of the models.}
\begin{tabular}{lr}
	\toprule
	Hyperparameter & Value \\
	\midrule
	Number of CNN layers in inference network & 1\\
	Number of filters per CNN layer & 256\\
	Filter size (i.e., time window size) & 3\\
	Number of feedforward layers in inference network & 2\\
	Width of feedforward layers & 256\\
	Dimensionality of latent space & 256\\
	Length scale of Cauchy kernel & 2.0\\
	Number of feedforward layers in generative network & 3\\
	Width of feedforward layers & 256\\
	Activation function of all layers & ReLU\\
	Learning rate during training & 0.001\\
	Optimizer & Adam \citep{kingma2014adam} \\
	Number of training epochs & 20\\
	Train/val/test split of data set & 50,000/10,000/10,000\\
	Dimensionality of time points & 784 \\
	Length of time series & 10 \\
	Tradeoff parameter $\beta$ & 0.8\\
	\bottomrule
\end{tabular}
\label{tab:hmnist_hyperparams}
\end{table*}

\subsection{SPRITES}

Since the SPRITES data also consist of images, we use the same 2D convolutional preprocessing as in the Healing MNIST experiment.
The hyperparameters are reported in Table~\ref{tab:sprites_hyperparams}.

\begin{table*}[h!]
\centering
\caption{Hyperparameters used in the GP-VAE model for the experiment on SPRITES. Some of the parameters are only relevant in a subset of the models.}
\begin{tabular}{lr}
	\toprule
	Hyperparameter & Value \\
	\midrule
	Number of CNN layers in inference network & 1\\
	Number of filters per CNN layer & 32\\
	Filter size (i.e., time window size) & 3\\
	Number of feedforward layers in inference network & 2\\
	Width of feedforward layers & 256\\
	Dimensionality of latent space & 256\\
	Length scale of Cauchy kernel & 2.0\\
	Number of feedforward layers in generative network & 3\\
	Width of feedforward layers & 256\\
	Activation function of all layers & ReLU\\
	Learning rate during training & 0.001\\
	Optimizer & Adam \citep{kingma2014adam} \\
	Number of training epochs & 20\\
	Train/val/test split of data set & 8,000/1,000/1,000\\
	Dimensionality of time points & 12288 \\
	Length of time series & 8 \\
	Tradeoff parameter $\beta$ & 0.1\\
	\bottomrule
\end{tabular}
\label{tab:sprites_hyperparams}
\end{table*}

\subsection{Real medical time series data}

For the Physionet 2012 data, we omitted the convolutional preprocessing, since these are not image data.
We hence directly feed the data into the 1D convolutional layer over time.
In this experiment, we follow the evaluation protocol from BRITS \citep{cao2018brits} and use the same set for training and evaluation. We randomly eliminate $10~\%$ of observed measurements from the data for validation.

The hyperparameters for this experiment are reported in Table~\ref{tab:physionet_hyperparams}.

\begin{table*}[h!]
\centering
\caption{Hyperparameters used in the GP-VAE model for the experiment on medical time series from the Physionet data set. Some of the parameters are only relevant in a subset of the models.}
\begin{tabular}{lr}
	\toprule
	Hyperparameter & Value \\
	\midrule
	Number of CNN layers in inference network & 1\\
	Number of filters per CNN layer & 128\\
	Filter size (i.e., time window size) & 24\\
	Number of feedforward layers in inference network & 1\\
	Width of feedforward layers & 128\\
	Dimensionality of latent space & 35\\
	Length scale of Cauchy kernel & 7.0\\
	Number of feedforward layers in generative network & 2\\
	Width of feedforward layers & 256\\
	Activation function of all layers & ReLU\\
	Learning rate during training & 0.001\\
	Optimizer & Adam \citep{kingma2014adam}\\
	Number of training epochs & 40\\
	Train/val/test split of data set & 4,000\\
	Dimensionality of time points & 35 \\
	Length of time series & 48 \\
	Tradeoff parameter $\beta$ & 0.2\\
	\bottomrule
\end{tabular}
\label{tab:physionet_hyperparams}
\end{table*}

\newpage

\section{Additional experiments}

\subsection{Missingness rates on Healing MNIST}

To explore the influence of the missingness rates on the performance of the models, we conducted an experiment where we introduced missingness into the Healing MNIST time series at different rates between 10~\% and 90~\%.
We compare the performance in terms of negative log likelihood (\ref{tab:hmnist_rates_likelihood}).

\begin{table*}[h!]
\centering
\caption{Performance of different models on Healing MNIST data with artificial missingness and different missingness rates. We report negative log likelihood (lower is better). The reported values are means and their respective standard errors over the test set.}
\resizebox{\linewidth}{!}{
\begin{tabular}{lrrrr}
\toprule
Missingness & VAE \citep{kingma2013auto} & HI-VAE \citep{nazabal2018handling} & GP-VAE (RBF kernel) & GP-VAE (proposed) \\
\midrule
10 \% &  0.0125 $\pm$ 0.0000 &  0.0117 $\pm$ 0.0000  &  0.0113 $\pm$ 0.0000 &  \textbf{0.0109 $\pm$ 0.0000}\\
20 \% &  0.0319 $\pm$ 0.0001 &  0.0275 $\pm$ 0.0001  &  0.0262 $\pm$ 0.0001 &  \textbf{0.0258 $\pm$ 0.0001}\\
30 \% &  0.0640 $\pm$ 0.0002 &  0.0507 $\pm$ 0.0002  &  0.0478 $\pm$ 0.0001 &  \textbf{0.0465 $\pm$ 0.0001}\\
40 \% &  0.1106 $\pm$ 0.0003 &  0.0803 $\pm$ 0.0003  &  0.0781 $\pm$ 0.0002 &  \textbf{0.0743 $\pm$ 0.0002}\\
50 \% &  0.1931 $\pm$ 0.0006 &  0.1349 $\pm$ 0.0005  &  0.1217 $\pm$ 0.0004 &  \textbf{0.1196 $\pm$ 0.0004}\\
60 \% &  0.3360 $\pm$ 0.0010 &  0.2218 $\pm$ 0.0008  &  0.1982 $\pm$ 0.0006 &  \textbf{0.1957 $\pm$ 0.0006}\\
70 \% &  0.6216 $\pm$ 0.0019 &  0.3818 $\pm$ 0.0014  &  0.3426 $\pm$ 0.0010 &  \textbf{0.3261 $\pm$ 0.0010}\\
80 \% &  1.3284 $\pm$ 0.0042 &  0.7613 $\pm$ 0.0025  &  0.6798 $\pm$ 0.0020 &  \textbf{0.6488 $\pm$ 0.0019}\\
90 \% &  4.0610 $\pm$ 0.0127 &  2.1878 $\pm$ 0.0064  &  1.9424 $\pm$ 0.0054 &  \textbf{1.8464 $\pm$ 0.0050}\\
\bottomrule
\end{tabular}
}
\label{tab:hmnist_rates_likelihood}
\end{table*}

It can be seen that the proposed model outperforms the other deep architectures (including the GP-VAE with an RBF kernel) for all the different missingness rates.
This also highlights that the Cauchy kernel does indeed help in modeling the temporal dynamics.

\subsection{Missingness mechanisms on Healing MNIST}
\label{sec:missingness_mechanisms}

So far, in our synthetic experiments we only looked at artificial missingness that was introduced completely at random (MCAR), that is, we uniformly sampled features to be missing independently of each other and their value.
In this experiment, we explore a few more structured missingness mechanisms which are described in the following.
The average missingness rate for all the different mechanisms is around 50~\%.

\paragraph{Feature correlation (\emph{Spatial}).}
We assume different features to be correlated in their missingness, that is, a feature is more likely to be missing if certain other features are missing.
We implement that by defining a spatial Gaussian process with RBF kernel on the Healing MNIST images and drawing the missingness patterns as samples from this process.
Neighboring pixels are therefore correlated in their missingness.

\paragraph{Positive temporal correlation (\emph{Temporal$^+$}).}
The missingness of features is positively correlated in time, that is, if a feature is missing at one time step, it is more likely to be missing at the consecutive time step.
We implement this again with a Gaussian process with RBF kernel, this time defined over time for each feature separately.

\paragraph{Negative temporal correlation (\emph{Temporal$^-$}).}
The missingness of features is negatively correlated in time, that is, if a feature is missing at one time step, it is less likely to be missing at the consecutive time step.
We implement this with a determinantal point process (DPP) over time for each feature separately.

\paragraph{Missingness not at random (\emph{MNAR}).}
In this setting, the missingness is actually dependent on the underlying ground-truth value of the feature.
In our example, white pixels in the Healing MNIST images are twice as likely to be missing than black pixels.

We assessed our model and the baselines on all these different settings and report the results in terms of likelihood in Table~\ref{tab:hmnist_mechanisms_likelihood}.

\begin{table*}[h!]
\centering
\caption{Performance of different models on Healing MNIST data with artificial missingness and different missingness mechanisms. We report negative log likelihood (lower is better). The reported values are means and their respective standard errors over the test set.}
\begin{tabular}{lrrr}
\toprule
Mechanism &   VAE \citep{kingma2013auto} & HI-VAE \citep{nazabal2018handling} &            GP-VAE (proposed) \\
\midrule
Spatial &  0.4802 $\pm$ 0.0016 &  0.2259 $\pm$ 0.0010 &  \textbf{0.1779 $\pm$ 0.0007} \\
Temporal$^+$ &  0.1918 $\pm$ 0.0006 &  0.1332 $\pm$ 0.0005 &  \textbf{0.1206 $\pm$ 0.0004} \\
Temporal$^-$ &  0.1940 $\pm$ 0.0006 &  0.1349 $\pm$ 0.0005 &  \textbf{0.1175 $\pm$ 0.0004} \\
MNAR & 0.4798 $\pm$ 0.0016 & 0.2896 $\pm$ 0.0010 & \textbf{0.2606 $\pm$ 0.0008} \\
\bottomrule
\end{tabular}
\label{tab:hmnist_mechanisms_likelihood}
\end{table*}

We observe that our proposed model outperforms all baselines in terms of likelihood and MSE on all the different missingness mechanisms.
We also observe that the MNAR setting seems to be the hardest one for the VAE-based models, followed by the setting with correlated features, whereas the settings with temporal correlation do not seem to be harder than the completely random ones (compare to Tab.~\ref{tab:hmnist_rates_likelihood}).
For the single imputation methods (mean and forward imputation), the MNAR setting also seems to be the hardest one, while the correlated features are not much harder than random.
However, the positive temporal correlation is harder for those methods and the negative temporal correlation is even easier than random missingness.

\end{document}